\pdfoutput=1
\documentclass[sigconf, nonacm]{acmart}

\usepackage{booktabs}
\usepackage{graphicx}
\usepackage{amsmath}
\usepackage{etoolbox}

\definecolor{headblue}{HTML}{1F3A68}
\DeclareRobustCommand\eqcontribmark{%
  \textcolor{headblue}{\scalebox{1.25}{$\bigstar$}}}
\newcommand\preprintmark{%
  \textcolor{headblue}{\sffamily\bfseries\footnotesize PREPRINT}}
\makeatletter
\patchcmd{\@mktitle@i}{\@title}{\textcolor{headblue}{\@title}}{}{}
\patchcmd{\@mktitle@iii}{\@title}{\textcolor{headblue}{\@title}}{}{}
\patchcmd{\@mktitle@iv}{\@title}{\textcolor{headblue}{\@title}}{}{}
\preto\@secfont{\color{headblue}}
\preto\@subsecfont{\color{headblue}}
\preto\@subsubsecfont{\color{headblue}}
\let\preprint@fnsymbol\@fnsymbol
\def\@fnsymbol#1{\ifnum#1=\@ne\eqcontribmark\else\preprint@fnsymbol{#1}\fi}
\AtBeginDocument{%
  \let\preprint@standard\ps@standardpagestyle
  \def\ps@standardpagestyle{\preprint@standard
    \fancyhead[RO,LE]{\preprintmark}}%
  \let\preprint@first\ps@firstpagestyle
  \def\ps@firstpagestyle{\preprint@first
    \fancyhead[RO,LE]{\preprintmark}}%
  \pagestyle{standardpagestyle}}
\makeatother

\begin{document}
\title{Reading the Mask: Descriptor-Grounded BI-RADS Rationale Generation for
  Breast MRI Segmentation}

\author{Aarti Sharma}
\authornote{Both authors contributed equally to this research.}
\affiliation{%
  \institution{Department of Applied Mathematics and Scientific Computing,
    Indian Institute of Technology Roorkee}
  \city{Roorkee}
  \country{India}
}

\author{Hrishikesh Singh}
\authornotemark[1]
\affiliation{%
  \institution{Department of Applied Mathematics and Scientific Computing,
    Indian Institute of Technology Roorkee}
  \city{Roorkee}
  \country{India}
}

\author{Divyanshu Kumar}
\affiliation{%
  \institution{Indian Institute of Technology Madras}
  \city{Chennai}
  \country{India}
}

\author{Millie Pant}
\affiliation{%
  \institution{Department of Applied Mathematics and Scientific Computing,
    Indian Institute of Technology Roorkee}
  \city{Roorkee}
  \country{India}
}
\additionalaffiliation{%
  \institution{Mehta Family School of Data Science and Artificial Intelligence,
    Indian Institute of Technology Roorkee}
  \city{Roorkee}
  \country{India}
}

\renewcommand{\shortauthors}{Sharma and Singh, et al.}

\begin{abstract}
Deep learning based models accurately delineate tumours on dynamic
contrast-enhanced MRI (DCE-MRI), but segmentation masks alone do not provide
the rationales or reasoning used by radiologists to assess the lesion
suspicion. This limits the explainability as well the clinical integration. We
investigate whether clinically relevant descriptors can instead be derived from
the segmentation masks and their dynamic contrast-enhancement series. We have
used Breast Imaging Reporting and Data System (BI-RADS) guidelines as a
structured framework. Using these guidelines we designed a BI-RADS decision
tree with eight quantitative measurements using radiomics, together with the
lesion size and component count, that determine the descriptors at eight
nodes. Using these descriptors we generate
natural-language rationale using a large language model. The model was
constrained to express only the derived records from the decision tree. After
applying this framework to the expert and automatic masks of each of the 1,506
cases in the MAMA-MIA dataset, we generated a reproducible corpus of 3,012
BI-RADS grounded rationales. We obtained a
faithfulness score of 99.98\% and a counterfactual accuracy of 99.3\%,
indicating that the generated rationales faithfully verbalises the measured
descriptors and correctly reflects changes in the underlying record. Agreement
between the descriptors from expert and automatic masks falls from 0.832 to
0.549 as mask quality falls from Good to Missed. The generated rationales are
intended as draft
representations for radiologist review and final sign-off rather than
autonomous clinical reports.
\end{abstract}



\maketitle


\section{Introduction}

When radiologists analyse the contrast-enhanced breast MRI, they use
standardised guidelines such as BI-RADS (Breast Imaging-Reporting and Data
System) to locate lesions, describe their morphology and assess malignancy
risk. On the other hand, the automated segmentation methods tell only a lesion
boundary or mask. They lack explanation of the image features which support the
results. This lack of clinically meaningful rationales makes the automated
segmentation harder to interpret and trust. This creates a barrier in their
clinical adoption.

There are several methods which improve interpretability. Post-hoc methods such
as Grad-CAM~\cite{gradcam} identify the influential image regions but do not
associate them with clinical descriptors. More recently, the researchers
in~\cite{prostate} have generated PI-RADS style reasoning alongside lesion
masks but their methodology relies on radiologist-authored reports. Such
report-based supervision is expensive and difficult to obtain for large
retrospective datasets.

We propose a report-independent framework for generating BI-RADS grounded
rationales from breast DCE-MRI. Our key observation is that many breast BI-RADS
descriptors such as shape, margin, internal enhancement and kinetic behaviour
- that is, how the lesion's contrast enhancement changes over time after the
contrast injection - can be quantitatively derived from the segmentations and
dynamic MRI data. Thus, we reverse the conventional pipeline. We compute these
measurements directly from images and masks, then we convert the resulting
measurements into structured BI-RADS descriptors and natural-language
rationales. We apply the framework to all 1,506 cases in
MAMA-MIA~\cite{mamamia} without additional radiologist annotation and evaluate
the generated rationales using four annotation-free metrics: faithfulness,
hallucination rate, omission rate, and counterfactual accuracy. Since every
case has both an expert and an automatic mask, we also measure how much the
descriptors change when the automatic mask is used. This provides a
scalable alternative to report-dependent approaches for producing clinically
grounded explanations of breast MRI lesions.

\section{Related Work}

Encoder-decoder networks~\cite{unet} remain the standard for volumetric lesion
delineation, with nnU-Net~\cite{nnunet} providing a self-configuring pipeline which is
challenging to outperform without task-specific optimisation. Segmentation
models adapted to medical data, MedSAM~\cite{medsam}, extend promptable
segmentation to radiological volumes. But these methods do not provide
interpretable outputs beyond the mask itself. Another class of interpretable
methods involve gradient-based methods~\cite{gradcam}, which assign importance
to pixels but leave the clinical rationale of the highlighted region
unspecified. Concept bottleneck models~\cite{cbm} provide named intermediates,
but further analysis~\cite{cbmintended} has questioned whether the learned
concepts correspond to what their names suggest. Prototype-based
classifiers~\cite{protopnet} use a representative example of a particular class
to develop reasoning, but they are limited by the set of prototypes they
learned beforehand. Contrastive image-text pretraining~\cite{clip}, adapted to
biomedical corpora as in BiomedCLIP~\cite{biomedclip}, aligns the whole image
with entire captions. This type of global alignment cannot distinguish multiple
findings in the same organ.

Quantitative DCE-MRI features have been related to BI-RADS
descriptors~\cite{nie2008,gorji2025}, kinetic curves classified
automatically~\cite{chen2006}, and template-based reports generated for chest
X-rays~\cite{pino2021}. We add the record as the only input to a text
generator, and a paired comparison of expert and automatic masks.

\section{Dataset}

MAMA-MIA~\cite{mamamia} is a multi-centre collection of 1,506 breast DCE-MRI
studies assembled from four public TCIA~\cite{tcia} cohorts:
ISPY2~\cite{hylton}, DUKE~\cite{saha}, ISPY1~\cite{hylton} and NACT. This
dataset contains expert masks and autogenerated masks. The autogenerated masks
are graded by quality by two experts visually. The released grades are Good,
Acceptable, Poor and Missed. There are three to six post-contrast phases per
study. Acquisition timestamps are provided for each study. A multifocality
indicator has also been provided for 1,346 cases.

%

\section{Proposed Approach}

\begin{figure*}[t]
  \centering
  \includegraphics[alt={Three-panel overview of the pipeline. Panel 1, input data: the pre-contrast, early and late phases of a breast DCE-MRI series, and two tumour outlines for the same case, an expert segmentation and an automatic segmentation. Panel 2, measurement and rule lookup: the largest lesion is kept, shape, edge, dye-pattern and dye-timing measurements are computed, and a BI-RADS decision tree turns them into descriptors. Panel 3, draft description: for each measurement family an example image and the evidence behind the term, for example an outline far from ball-like gives an irregular shape and brightness that falls after the early scan gives washout, ending in the sentence: An irregular mass with spiculated margins and rim enhancement; contrast washes out in the delayed phase.},
    width=\textwidth]{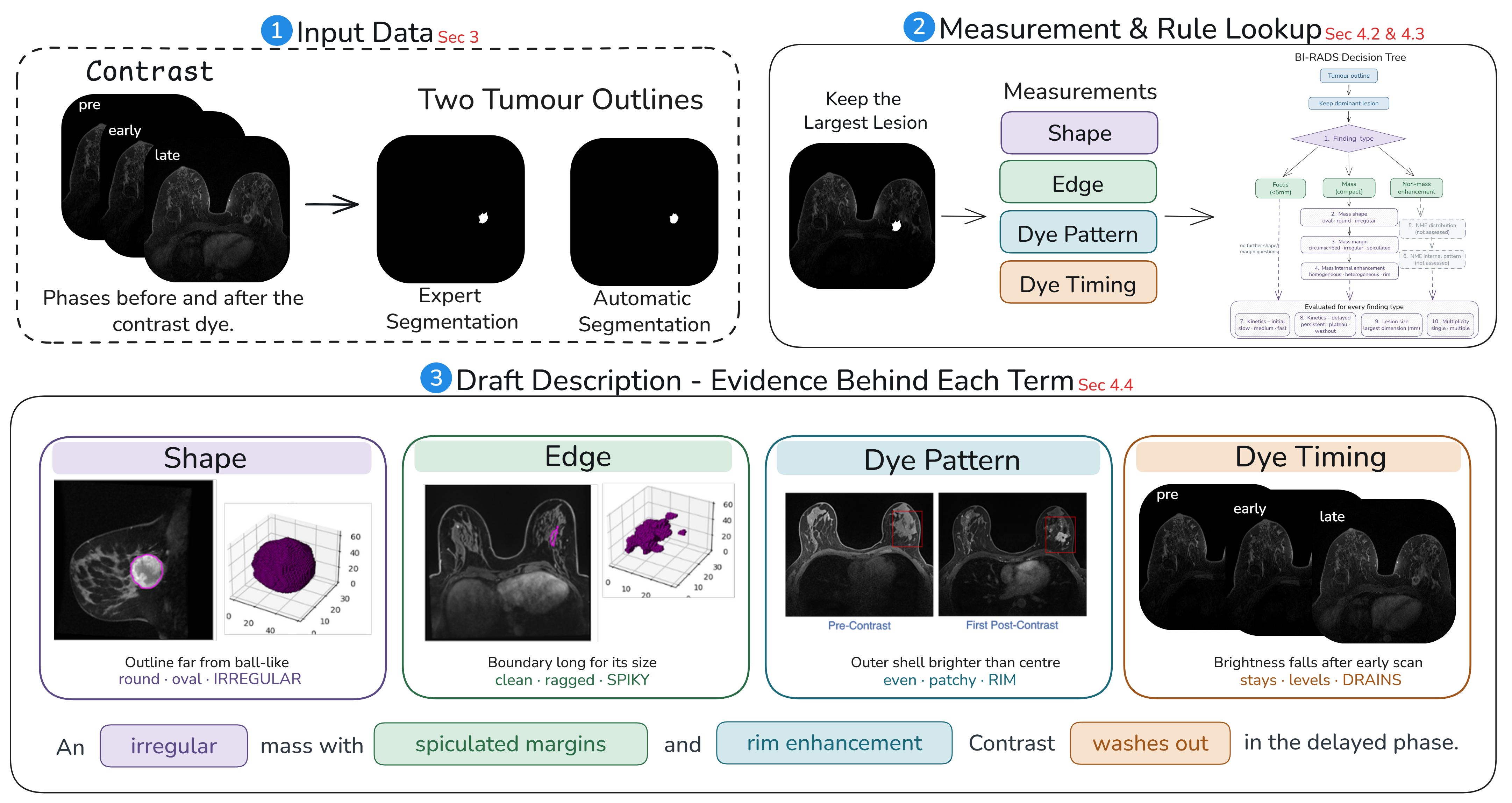}
  \caption{Overall pipeline of the proposed methodology.}
  \label{fig:pipeline}
\end{figure*}

Our method generates rationales grounded in the ACR BI-RADS Atlas (5th edition)
MRI lexicon~\cite{birads} for the breast DCE-MRI dataset consisting of lesion
segmentation masks. First, we prepare the lesion representation
(Section~\ref{sec:lesion-rep}), then we extract the quantitative lesion
measurements (Section~\ref{sec:measurements}). These features describe the
morphology as well as the contrast-enhancement behaviour of the lesions. We
then map these measurements to the BI-RADS descriptors
(Section~\ref{sec:descriptors}), finally a large language model (LLM) converts
the structured BI-RADS records into a natural-language rationale
(Section~\ref{sec:generation}).

\subsection{Step 1: Preparing the Lesion Representation}
\label{sec:lesion-rep}

The segmentation masks may contain multiple disconnected regions, which is
common in multifocal disease. Computing the shape features on the full mask may
therefore give misleading values, hence we first identify the connected
components and then select the dominant lesion for morphological analysis.

\subsubsection{Connected component analysis}

Let $M$ denote the binary lesion mask. We identify its connected components
using 26-connec\-tiv\-ity:
\begin{equation}
  M = \bigcup_{i=1}^{m} C_i ,
\end{equation}
where $m$ is the total number of connected components. Two voxels are
considered connected if they share a face, edge, or corner.

We define the dominant component as the largest connected component:
\begin{equation}
  M^{*} = \arg\max_{C_i} |C_i| .
\end{equation}

All subsequent morphological measurements are computed on $M^{*}$. We retain
$m$ as the multiplicity descriptor: \emph{single} if $m = 1$ and
\emph{multiple} otherwise. No minimum component size is applied.

We also compute the dominance ratio:
\begin{equation}
  \rho = \frac{|M^{*}|}{|M|} .
\end{equation}

A high $\rho$ indicates that most of the segmented lesion belongs to one
component. A low $\rho$ indicates that the segmentation is spread across
several components. We retain this value as a diagnostic measure of mask
dispersion.

\subsubsection{Spatial normalisation}

The MRI volumes have different voxel spacings across cases. We therefore
resample the lesion region to isotropic $1\,\mathrm{mm}^{3}$ spacing before
computing surface-based features. Isotropic resampling is important because
surface area and volume depend on voxel spacing. Without normalisation,
measurements from different cases would not be directly comparable.

\subsection{Step 2: Extracting Quantitative Lesion Measurements}
\label{sec:measurements}

We extract quantitative measurements from the dominant lesion component
$M^{*}$. The measurements capture two aspects of the lesion: its morphology and
its contrast-enhancement behaviour. Morphological measurements describe the
lesion shape and boundary. The enhancement measurements describe the spatial
and temporal distribution of contrast uptake.

\subsubsection{Morphological measurements}

Let $V$ and $A$ denote the volume and surface area of $M^{*}$. Let
$\lambda_1 \geq \lambda_2 \geq \lambda_3$ denote the eigenvalues of the
covariance matrix of the voxel coordinates of $M^{*}$. We compute four
morphological measurements.

\emph{Sphericity} measures how closely the lesion resembles a sphere:
\begin{equation}
  \varphi_1 = \frac{(36\pi V^{2})^{1/3}}{A} .
\end{equation}
A spherical lesion has a sphericity close to one. Irregular lesions have lower
values.

\emph{Elongation} measures the relative extent of the lesion along its
principal axes:
\begin{equation}
  \varphi_2 = \sqrt{\frac{\lambda_3}{\lambda_1}} .
\end{equation}
A lower value indicates a more elongated lesion, while a value closer to one
indicates a more isotropic shape.

\emph{Surface-to-volume ratio} measures the amount of boundary relative to
lesion volume:
\begin{equation}
  \varphi_3 = \frac{A}{V} .
\end{equation}
Irregular or complex boundaries have a higher surface-to-volume ratio. We
therefore use this measurement as a quantitative proxy for the BI-RADS margin
descriptor. Note that $\varphi_3$ also depends on size ($3/r$ for a sphere).

\emph{Compactness} measures how much of the lesion's bounding box is occupied
by the lesion:
\begin{equation}
  \varphi_4 = \frac{|M^{*}|\, v}{V_{\mathrm{bbox}}} ,
\end{equation}
where $v$ is the voxel volume and $V_{\mathrm{bbox}}$ is the volume of the
axis-aligned bounding box. This measurement helps distinguish compact,
spatially concentrated enhancement from more dispersed regions.

Together, these four measurements provide a quantitative description of lesion
shape and boundary structure.

\subsubsection{Contrast-enhancement measurements}

We next measure how the lesion enhances after contrast administration. Let
$I_0$ denote the pre-contrast image, $I_{t_e}$ the first post-contrast image,
and $I_{t_\ell}$ the final post-contrast image. These phases are selected using
the acquisition timestamps provided for each case.

We first compute the early enhancement map:
\begin{equation}
  E = I_{t_e} - I_0 .
\end{equation}
For any region $S$, let $\mathbb{E}_{S}[\cdot]$ denote the mean intensity over
its voxels.

To measure peripheral enhancement, we divide the lesion into a core and a
boundary shell. The core is obtained by eroding $M^{*}$ three times with the
six-connected structuring element:
\begin{equation}
  M^{*}_{\mathrm{core}} = \mathrm{erode}(M^{*},\, 3\ \text{iterations}) ,
\end{equation}
and the shell is
\begin{equation}
  M^{*}_{\mathrm{shell}} = M^{*} \setminus M^{*}_{\mathrm{core}} .
\end{equation}

The rim ratio is then defined as
\begin{equation}
  \varphi_5 =
  \frac{\mathbb{E}_{M^{*}_{\mathrm{shell}}}[E]}
       {\mathbb{E}_{M^{*}_{\mathrm{core}}}[E]} .
\end{equation}
A higher value indicates stronger enhancement near the lesion boundary relative
to its core. This provides a quantitative measure for the rim-enhancement
descriptor. If the eroded core is too small, $\varphi_5$ is not computed and
internal enhancement is marked as not assessed ($\perp$).

We measure enhancement heterogeneity using the standard deviation of
enhancement within the lesion. The intensities are not normalised across
scanners:
\begin{equation}
  \varphi_6 =
  \sqrt{\mathbb{E}_{M^{*}}\!\left[\left(E-\mathbb{E}_{M^{*}}[E]\right)^{2}\right]} .
\end{equation}
Higher values indicate greater variation in enhancement across the lesion.

To capture temporal enhancement behaviour, we compute the signal enhancement
ratio (SER):
\begin{equation}
  \varphi_7 =
  \frac{\mathbb{E}_{M^{*}}[I_{t_e} - I_0]}
       {\mathbb{E}_{M^{*}}[I_{t_\ell} - I_0]} .
\end{equation}
A higher SER indicates that enhancement is stronger in the early phase than in
the late phase, providing a numerical measure of washout behaviour. More
precisely, $\mathrm{SER} > 1$ means early enhancement exceeds late enhancement,
so the signal has decreased from early to late, indicating \emph{washout};
$\mathrm{SER} \approx 1$ means early and late enhancement are similar,
indicating an approximate \emph{plateau}; and $\mathrm{SER} < 1$ means late
enhancement is greater, indicating \emph{persistent} or progressive
enhancement.

Finally, we compute the percentage initial enhancement:
\begin{equation}
  \varphi_8 = 100 \times
  \frac{\mathbb{E}_{M^{*}}[E]}{\mathbb{E}_{M^{*}}[I_0]} .
\end{equation}
This quantity is used to determine the initial kinetic descriptor.

Thus, the measurement stage produces a compact set of eight proxy quantitative
values ($\varphi_1$--$\varphi_8$) that describe lesion morphology, internal
enhancement, and temporal enhancement. The decision tree also uses the maximum
3D diameter $d$ of $M^{*}$ and the component count $m$. These values are used
in the next stage to assign the corresponding BI-RADS descriptors.

\subsection{Step 3: Assigning BI-RADS Descriptors}
\label{sec:descriptors}

The quantitative measurements are continuous values, while BI-RADS uses
categorical descriptors. We therefore map the measurements to BI-RADS values
using predefined decision rules. We use the BI-RADS MRI lexicon from the ACR
BI-RADS Atlas (5th edition)~\cite{birads} and retain all 14 nodes in its
reporting sequence.

\subsubsection{BI-RADS decision tree}

Of the 14 nodes, eight are computed from the MRI and lesion mask: finding type,
mass shape, mass margin, mass internal enhancement, initial kinetic phase,
delayed kinetic phase, lesion size, and multiplicity.

The remaining six nodes are not computed and are assigned $\perp$. Non-mass
distribution and internal enhancement pattern require ductal-anatomy
information that is not available from the lesion mask. Associated features,
breast composition, and background parenchymal enhancement require information
outside the lesion mask. The final assessment category is also excluded because
it requires clinical history and prior imaging. Moreover, all cases in MAMA-MIA
are biopsy-confirmed malignant, so assigning an assessment category would not
provide a meaningful evaluation.

We retain these nodes explicitly as $\perp$ rather than removing them. This
keeps the representation aligned with the complete BI-RADS sequence.

The decision tree also contains conditional nodes. In particular, mass shape,
mass margin, and mass internal enhancement are evaluated only when the finding
type is classified as a mass. This prevents descriptors that require a defined
mass boundary from being assigned to non-mass enhancement.

\begin{figure*}[t]
  \centering
  \includegraphics[alt={Decision tree with ten numbered nodes; each test compares one measurement with one threshold and leaves are BI-RADS descriptor values. Node 1, finding type on the dominant lesion: diameter below 5 mm gives Focus (no case in MAMA-MIA); otherwise compactness of at least 0.15 gives Mass, else non-mass enhancement, whose nodes 5 and 6 are not assessed. Evaluated only for a mass: node 2, shape (sphericity below 0.487 gives irregular, else elongation of at least 0.775 gives round, else oval); node 3, margin (surface-to-volume ratio below 0.363 gives circumscribed, below 0.508 irregular, else spiculated); node 4, internal enhancement (not assessed if the rim ratio is undefined, rim if it is at least 0.965, else heterogeneous if heterogeneity is at least 123.0, else homogeneous). Evaluated for every finding type: node 7, initial kinetics (enhancement below 50 percent slow, below 100 percent medium, else fast); node 8, delayed kinetics (signal enhancement ratio below 0.9 persistent, up to 1.1 plateau, else washout); node 9, lesion size in mm; node 10, multiplicity (one component single, else multiple). Nodes 11 to 14 are not assessed.},
    width=\textwidth]{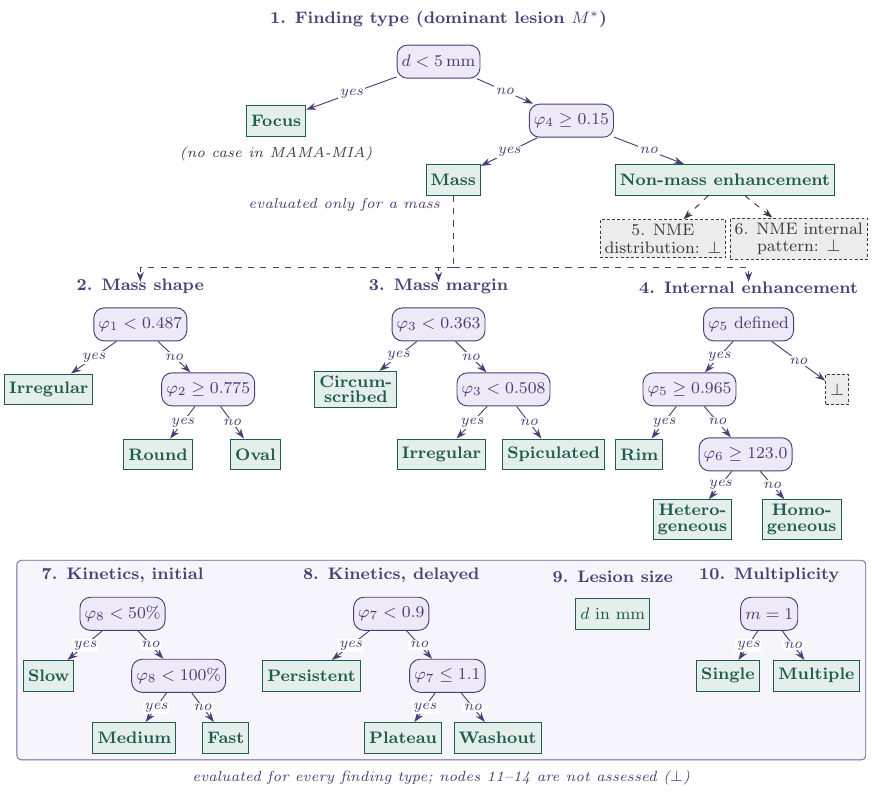}
  \caption{BI-RADS decision tree. Each test compares one measurement with its
    threshold; leaves are descriptor values.}
  \label{fig:decision-tree}
\end{figure*}

\subsubsection{Descriptor thresholds}

We use four types of threshold rules. Fig.~\ref{fig:decision-tree} shows every
test with its threshold; where two measurements decide one node, they are
tested in the order shown.

\emph{Standard-defined thresholds} are taken directly from the BI-RADS
definition. For example, a lesion with a maximum 3D diameter below 5\,mm is
classified as a focus rather than a mass. No lesion in MAMA-MIA falls in
this branch.

For measurements without an established external threshold, we derive
\emph{cohort-based thresholds} from the expert masks. For a measurement
$\varphi_j$, let $\hat{F}_j$ denote its empirical cumulative distribution
function. A threshold at quantile $q$ is
\begin{equation}
  \tau_j(q) = \hat{F}_j^{-1}(q).
\end{equation}
We use $q = 1/3$ and $2/3$ for surface-to-volume ratio, $q = 1/3$ for
sphericity, $q = 2/3$ for rim ratio, and the median for elongation and
heterogeneity. Quantile thresholds fix how often each descriptor value occurs
on the expert masks, so they place a lesion relative to the rest of this cohort
rather than against an absolute clinical criterion.

For the mass versus non-mass decision, we use a \emph{fixed} compactness
threshold of 0.15 rather than a cohort quantile. A quantile-based threshold
would force a fixed share of a biopsy-confirmed malignant cohort into non-mass
enhancement, which contradicts the clinical expectation. We have calibrated it
so that the resulting split is plausible (12.8\% non-mass enhancement on the
expert masks); it was not tuned against any reference label.

For kinetic descriptors, we use \emph{conventional thresholds} from prior
work~\cite{kuhl}. Initial enhancement is divided at 50\% and 100\%, giving
slow, medium, and fast enhancement. The SER thresholds are 0.9 and 1.1, giving
persistent, plateau, and washout patterns.

All cohort-derived thresholds are estimated once using the expert masks. The
same thresholds are then applied unchanged to the automatic masks. This
prevents the automatic-mask results from changing the descriptor definitions.
The thresholds are therefore fitted and applied on the same cohort.

\subsubsection{Constructing the structured BI-RADS record}

The assigned descriptors are collected into a structured BI-RADS record. Let
the record be
\begin{equation}
  \ell = \{\ell_1, \ell_2, \ldots, \ell_{14}\},
\end{equation}
where each $\ell_k$ corresponds to one node in the BI-RADS decision tree.

Each node contains either its assigned descriptor value or $\perp$ when the
node is not evaluated. Conditional nodes are evaluated only when their
preconditions are satisfied.

The resulting record contains the computed descriptors for finding type, shape,
margin, internal enhancement, kinetic behaviour, lesion size, and multiplicity.
The six unsupported nodes remain explicitly marked as $\perp$.

This record is the only representation passed to the language model. A
conditional field that is $\perp$ is passed as \emph{not assessed}, and lesion
size is not passed (it is compared separately, Sec.~\ref{sec:agreement}). The
original MRI, lesion mask, and continuous measurements are not provided to it.
This separates descriptor computation from language generation and allows the
generated rationale to be evaluated directly against the known input record.

\subsection{Step 4: Descriptor-Grounded Rationale Generation}
\label{sec:generation}

The structured BI-RADS record is converted into natural-language text using
Llama-3.3-70B. The model is used only as a verbalisation module.

The model receives three inputs: the BI-RADS decision-tree specification, the
structured record $\ell$, and ten hand-written examples. It is instructed
to describe only the information present in $\ell$ and not to give a
diagnostic verdict. It runs locally with Ollama at temperature 0.2, once for
each of the 352 distinct records. A fixed template is equally faithful
(Sec.~\ref{sec:ablation}); the model gives more natural phrasing, not
evaluated with readers.

The model does not receive the MRI images, lesion mask, or raw measurements.
Thus, the generation process is
\begin{equation}
  r = g(\ell),
\end{equation}
where $g$ denotes the LLM verbaliser and $r$ is the generated rationale.

Since the model receives only the structured descriptor record, it cannot
directly derive new observations from the MRI. This design keeps the rationale
tied to the descriptors produced by the measurement and quantisation stages.

\subsection{Step 5: Evaluating Rationale Fidelity}
\label{sec:fidelity}

We evaluate whether the generated rationale correctly expresses the structured
BI-RADS record provided to the LLM. Let $\ell$ denote the input record, $r$ the
generated rationale, and $\hat{\ell}$ the record recovered from $r$ by matching
the same closed descriptor lexicon. A \emph{not assessed} field is correct
when the text says that it does not apply. We use four metrics.

\emph{Faithfulness} measures the fraction of input fields that are correctly
expressed in the rationale:
\begin{equation}
  \mathrm{Faithfulness} =
  \frac{|\hat{\ell} \cap \ell_{\mathrm{correct}}|}{|\ell|} .
\end{equation}

\emph{Hallucination rate} measures the fraction of statements in the rationale
that are not supported by the input record:
\begin{equation}
  \mathrm{Hallucination\ rate} =
  \frac{N_{\mathrm{unsupported}}}{N_{\mathrm{total}}} ,
\end{equation}
where $N_{\mathrm{unsupported}}$ counts sizes, lateralities or locations,
which the record never supplies; it is a lower bound.

\emph{Omission rate} measures the fraction of input fields that are not
expressed in the rationale:
\begin{equation}
  \mathrm{Omission\ rate} = \frac{|\ell \setminus \hat{\ell}|}{|\ell|} .
\end{equation}

Finally, \emph{counterfactual accuracy} tests whether the rationale responds to
controlled changes in the input record. We deliberately change one
descriptor value in $\ell$, regenerate the rationale, and check whether the
change is correctly reflected and no other field changes. If $N_{\mathrm{flip}}$ fields are changed and
$N_{\mathrm{prop}}$ changes are correctly expressed, then
\begin{equation}
  \mathrm{Counterfactual\ accuracy} =
  \frac{N_{\mathrm{prop}}}{N_{\mathrm{flip}}} .
\end{equation}

These metrics measure four different failure modes: incorrect verbalisation,
unsupported content, missing information, and failure to respond to changes in
the structured record. All four check how well $r$ states $\ell$; they do
not check whether $\ell$ itself is clinically correct.

\section{Implementation Details}

The shape descriptors are computed using PyRadiomics 3.1.0 with B-spline
interpolation~\cite{pyradiomics}. Connectivity analysis and morphological
erosion use SciPy and volumes are read with SimpleITK. The language model is
served by Ollama 0.30.6 on a single NVIDIA RTX 5000 Ada GPU.


\section{Results}

The pipeline was applied to all 1,506 MAMA-MIA studies using both the expert
and the automatic segmentation masks of each case. We got a total of 3,012
symbolic records. For every case the descriptor extraction was completed. On
the expert masks 1,313 cases had the finding-type node as mass and 193 as
non-mass enhancements (1,150 and 356 on the automatic masks); no case was a
focus. The internal enhancement node was returned as not assessed in 35
expert-mask and 20 automatic-mask cases, where the eroded lesion core was too
small to calculate the rim measurement.

\subsection{Faithfulness}
\label{sec:faithfulness}

We parsed each of the 3,012 generated rationales back to the closed
descriptor lexicon. We recovered 21,080 of 21,084 fields and achieved a
faithfulness score of 99.98\%. Of these fields, 1,702 are not assessed (the
mass-only fields of the 549 non-mass findings, and the 55 guard cases); all
were stated as not applicable, and over the remaining fields faithfulness is
also 99.98\%. The hallucination and omission rates are both zero across all
the 3,012 cases: no rationale stated a size, laterality or anatomical location,
none of which is passed to the model. Moreover, it omitted no field that the
record supplied. Out of the 21,084 fields four were wrong. They come from one
generated sentence, shared by four cases with the same record. The record said
that the initial kinetic phase was medium, but the model gave the rationale
``Fast initial enhancement is not seen, instead medium initial enhancement
\ldots{} is observed''. The text is correct, but since our parser matches the
descriptor terms in order and does not handle negation, it returns the denied
value. These four cases are therefore parser errors rather than model errors;
of the 352 distinct rationales, 351 parse without error.

\begin{table}[tbp]
  \caption{Descriptor-level performance of the verbalisation stage, over
    3,012 rationales. $\perp$ is the number of fields that are not assessed;
    the rationale must say that they do not apply. Contradiction counts a field
    stated with a different admissible value; it reduces faithfulness while
    counting as neither omission nor hallucination. Hallucination rate is
    0.0\% and counterfactual accuracy 99.3\% ($n = 575$).}
  \label{tab:per-descriptor}
  \label{tab:metrics}
  \centering
  \footnotesize
  \setlength{\tabcolsep}{2.5pt}
  \begin{tabular}{@{}lrrrrr@{}}
    \toprule
    BI-RADS node & Fields & $\perp$ & Faithful.\ (\%) & Omission (\%) & Contradict.\ (\%) \\
    \midrule
    Finding type          &  3012 &    0 & 100.0 & 0.0 & 0.0 \\
    Shape                 &  3012 &  549 & 100.0 & 0.0 & 0.0 \\
    Margin                &  3012 &  549 & 100.0 & 0.0 & 0.0 \\
    Internal enhancement  &  3012 &  604 & 100.0 & 0.0 & 0.0 \\
    Kinetics (initial)    &  3012 &    0 & 99.9 & 0.0 & 0.1 \\
    Kinetics (delayed)    &  3012 &    0 & 100.0 & 0.0 & 0.0 \\
    Multiplicity          &  3012 &    0 & 100.0 & 0.0 & 0.0 \\
    \midrule
    All fields            & 21084 & 1702 & 99.98 & 0.0 & 0.02 \\
    \bottomrule
  \end{tabular}
\end{table}

\begin{figure*}[t]
  \centering
  \begin{minipage}[t]{0.32\textwidth}
    \centering
    \includegraphics[alt={Line plot of descriptor agreement alpha, on the vertical axis, against binned Dice overlap of the automatic mask with the expert mask on the horizontal axis, with a 95 percent confidence band. Alpha rises with Dice; the band is widest in the two lowest Dice bins, which hold 30 and 20 cases. A dashed horizontal line marks the shuffled control.},
    width=\textwidth]{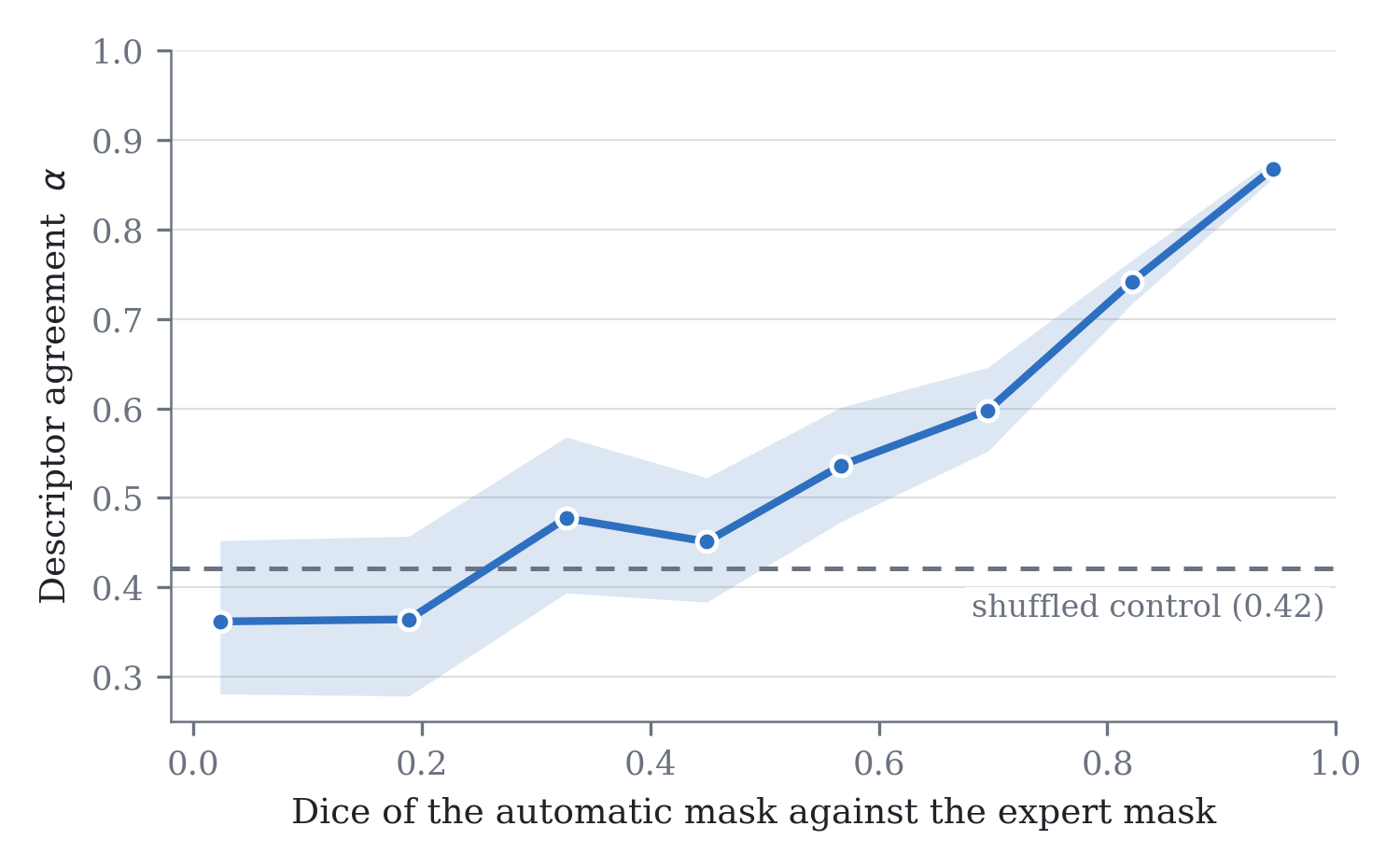}\\[2pt]
    {\footnotesize (a) $\alpha$ against Dice}
    \label{fig:alpha-dice}
  \end{minipage}
  \hfill
  \begin{minipage}[t]{0.32\textwidth}
    \centering
    \includegraphics[alt={Plot of descriptor agreement alpha by radiologist quality grade, from Good through Acceptable and Poor to Missed, for both graders with 95 percent confidence intervals. Alpha declines across the grades for both graders. A dashed horizontal line marks the shuffled control.},
    width=\textwidth]{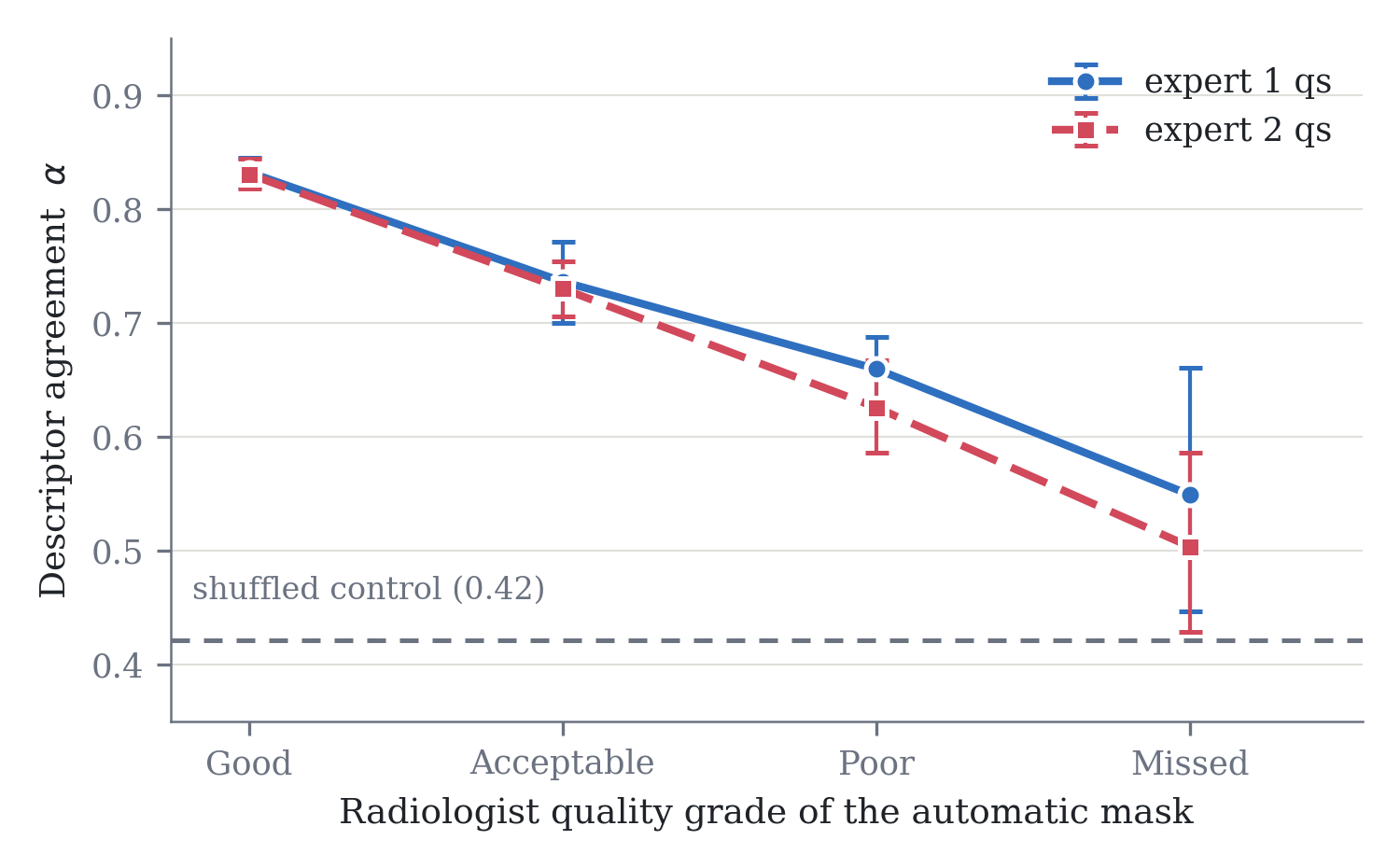}\\[2pt]
    {\footnotesize (b) $\alpha$ by quality grade}
    \label{fig:alpha-grade}
  \end{minipage}
  \hfill
  \begin{minipage}[t]{0.32\textwidth}
    \centering
    \includegraphics[alt={Per-node exact-match agreement between expert-mask and automatic-mask records across mask quality grades, one series per BI-RADS node, coloured by how the descriptor is measured. Delayed kinetics, averaged over the lesion, changes least as mask quality falls; the morphological descriptors, computed from the boundary, degrade sharply.},
    width=\textwidth]{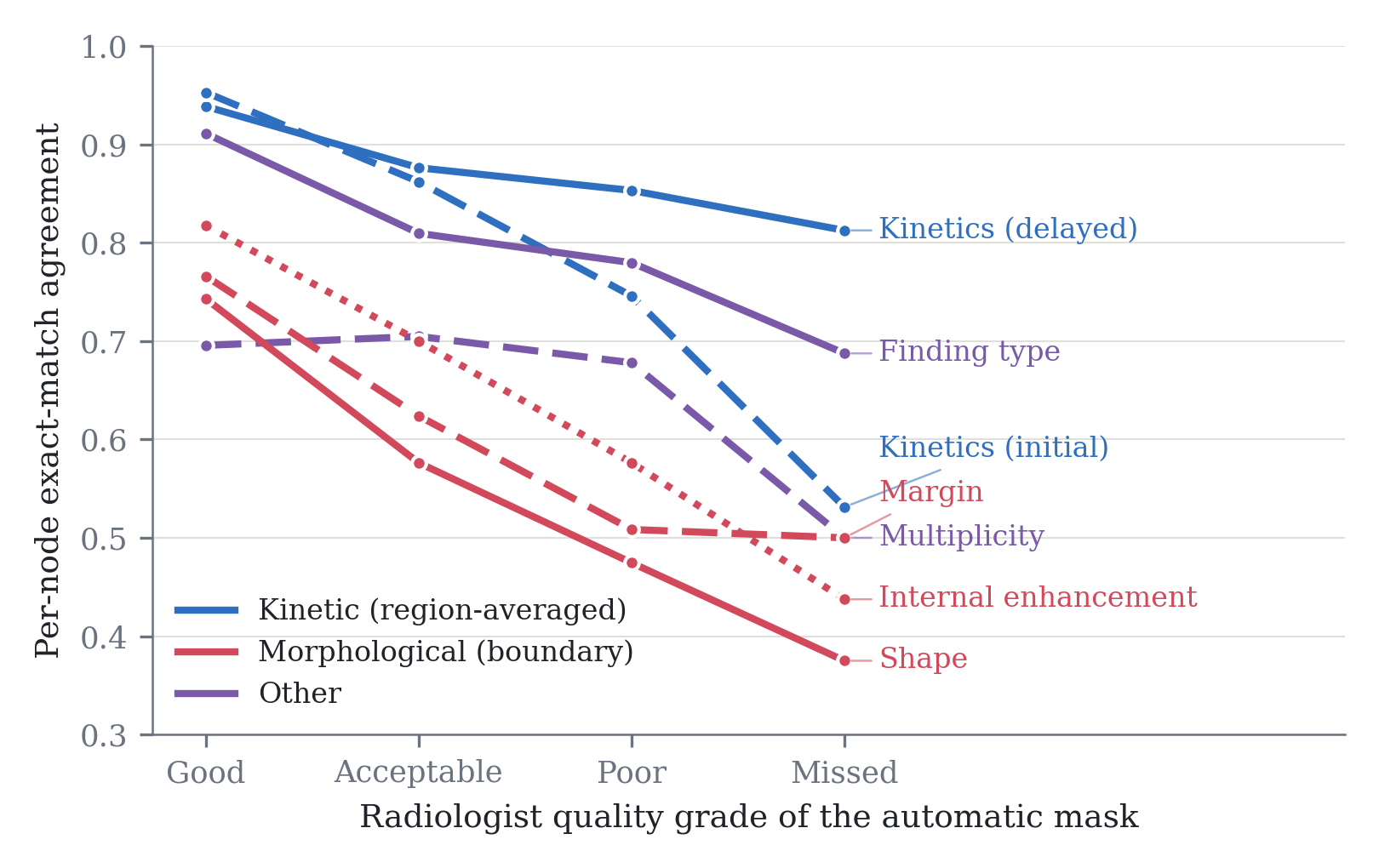}\\[2pt]
    {\footnotesize (c) per-node agreement}
    \label{fig:alpha-node}
  \end{minipage}

  \caption{Descriptor agreement under mask degradation.
    \textbf{(a)} Descriptor agreement $\alpha$ against the Dice overlap of the
    automatic mask with the expert mask, with a 95\% confidence band; the two
    lowest bins contain 30 and 20 cases respectively, which is why their bands
    are wide.
    \textbf{(b)} $\alpha$ by radiologist quality grade for both graders, with
    95\% confidence intervals. The second grader classified far fewer masks as
    Poor (158 versus 354) yet reproduces the same decline, so the trend follows
    mask quality rather than one rater's strictness.
    \textbf{(c)} Per-node exact-match agreement across mask quality grades,
    coloured by how each descriptor is measured: delayed kinetics, averaged over
    the lesion, is the least sensitive to boundary error, while morphological
    descriptors are computed from the boundary itself and degrade sharply.
    In (a) and (b) the dashed line is the shuffled control.}
  \label{fig:alpha}
\end{figure*}

\subsection{Counterfactual Consistency}

Counterfactual consistency was measured over 575 single-field perturbations of
100 records. We obtained a propagation rate of 1.000, meaning every
perturbation propagated into the regenerated text. In 4 of the 575 cases
(0.7\%) the parser also read a different value for a second, unperturbed
field. Therefore the counterfactual accuracy, which requires both that the
change appears and that nothing else moves, is 0.993. All four are
perturbations of the same record, whose rationale is the negated sentence
above; each regenerated text states the correct value, so this is the same
parser error.

\subsection{Descriptor Agreement under Mask Degradation}
\label{sec:agreement}

For each patient we have two records: one from the expert mask and the other
from the automatic mask. We calculate the descriptor agreement $\alpha$, which
tells how often the two agree, averaged over the seven categorical
descriptors. For a patient with
expert record $\ell^e$ and automatic record $\ell^a$, $\alpha$ is calculated as
\begin{equation}
  \alpha(\ell^e, \ell^a) =
  \frac{1}{|\mathcal{N}_c|}
  \sum_{k \in \mathcal{N}_c} \mathbb{1}\!\left[\ell^e_k = \ell^a_k\right],
  \label{eq:alpha}
\end{equation}
where $\mathbb{1}[\cdot]$ is 1 when the two values are identical and 0
otherwise. $\alpha$ compares descriptors; as the rationales state their
records almost exactly, it also describes the text. Lesion size is not part of
$\alpha$; its Pearson correlation between the two masks is 0.821. Across all
1,506 patients
$\alpha$ is 0.772. As a reference point,
pairing each expert record with a different patient's automatic record gives
0.422, so 0.422 is what chance agreement looks like.

The $\alpha$ score was 0.832 when masks were graded Good, 0.736 for Acceptable,
0.659 for Poor and 0.549 for Missed. The per-descriptor breakdown shows where
the loss actually occurred, which follows directly from how each descriptor is
measured. The kinetic descriptors are averages of signal intensity over the
whole lesion, so moving the boundary a little barely changes them: agreement
for delayed kinetics stays at 0.938 on Good masks and is still 0.812 on Missed
ones, although initial kinetics falls from 0.953 to 0.531. The
morphological descriptors are the boundary - shape, margin and internal
enhancement are computed from its geometry - so when the boundary is wrong
they fail. Shape falls from 0.743 to 0.375 and margin from 0.766 to 0.500 over
the same range.

Cohen's $\kappa$, which corrects for uneven label frequencies, shows the same
decline (Table~\ref{tab:alpha}); for delayed kinetics on Missed masks it is
0.415, and multiplicity has the lowest $\kappa$ (0.236).

\begin{table}[tbp]
  \caption{Descriptor agreement $\alpha$ (Eq.~\ref{eq:alpha}) and mean Cohen's
    $\kappa$ over the seven nodes, stratified by segmentation quality grade
    (grader 1). $n$ and Dice are properties of the dataset; $\alpha$ and
    $\kappa$ are measured here. The shuffled control pairs each expert
    record with a random other patient's automatic record.}
  \label{tab:alpha}
  \centering
  \small
  \begin{tabular}{@{}lrrrr@{}}
    \toprule
    Mask quality & $n$ & Mean Dice & $\alpha$ (95\% CI) & Mean $\kappa$ \\
    \midrule
    Good & 910 & 0.905 & 0.832 (0.819--0.845) & 0.686 \\
    Acceptable & 210 & 0.805 & 0.736 (0.700--0.771) & 0.528 \\
    Poor & 354 & 0.690 & 0.659 (0.631--0.688) & 0.425 \\
    Missed & 32 & 0.326 & 0.549 (0.446--0.661) & 0.206 \\
    \midrule
    Shuffled control & --- & --- & 0.422 & --- \\
    \bottomrule
  \end{tabular}
\end{table}

\subsection{Ablation Studies}
\label{sec:ablation}

We removed one component of the prompt at a time and regenerated 150 records
under each condition. Two reference conditions bound the range. The floor
shuffles the descriptor values at random before verbalisation, so the record is
clinically meaningless. The ceiling is a fixed template that fills the record
into a sentence with no language model at all. Conditions ran on the
Llama-3.3-70B language model, except the raw-measurement condition, which ran
on Qwen2.5-7B-Instruct. On Llama-3.3-70B the floor also reaches a faithfulness
of 1.000, so faithfulness shows how well a record is stated, not whether it is
right.

We identified the two conditions which constrain the model. The first is
removing the hand-written exemplars, which reduced the faithfulness to 0.728
and raised the omission rate to 0.270. In the second case, supplying the raw
continuous measurements in place of the descriptor words reduced the
faithfulness to 0.066, and omissions rose to 0.847 and hallucinations to 0.481.
The model begins to issue diagnostic statements the pipeline is designed to
exclude. The mapping from measurements to a closed vocabulary kept the model
tied to its evidence; this guarantee comes from the representation.

\section{Conclusion}

In this study we have shown that using BI-RADS as a structured decision tree
and defining eight measurements over a mask and its contrast series gives us
the symbolic records over the 1,506 MAMA-MIA studies. Using language models on
these records gives us clinical prose without ever observing an image.
Applying this to the expert and automatic masks of all 1,506 studies yields a
reproducible set of 3,012 BI-RADS-grounded rationales. The faithfulness score
achieved on the dataset is 99.98\%. This shows that the rationales state their
records correctly, not that the records are clinically correct. We also
measured the descriptor agreement on both mask sources and found that it falls
with segmentation mask quality. The descriptor agreement falls from 0.832 on
masks graded Good to 0.549 on those graded Missed, and the decline remains
after correcting for chance. The rationales produced are
drafts for a reporting radiologist to review, amend and sign off. In conclusion,
we have shown that descriptors for a breast MRI report can be measured from
segmentation masks and contrast series with the help of the structure provided
by the published BI-RADS lexicon.

\subsubsection{Limitations}
The descriptors are not yet compared with expert reading or with the
dataset's multifocality indicator; all cases are malignant; and the thresholds
are fitted on one dataset. Several measurements are simple proxies (e.g.\
surface-to-volume ratio depends on size, heterogeneity is not normalised).

\section*{Disclosure of Interests}
The authors have no competing interests to declare that are relevant to the
content of this article.

\bibliographystyle{ACM-Reference-Format}
\bibliography{references}

\end{document}